\documentclass[conference,final,letterpaper]{IEEEtran}
\IEEEoverridecommandlockouts
\usepackage{cite}
\usepackage{amsmath,amssymb,amsfonts}
\usepackage{algorithm, algorithmic}
\usepackage{graphicx}
\graphicspath{{03_graphics/}}
\usepackage{subcaption}
\usepackage{textcomp}
\usepackage{xcolor}
\usepackage{url}
\usepackage{definitions}
\usepackage[colorinlistoftodos, english, disable]{todonotes} 
\def\BibTeX{{\rm B\kern-.05em{\sc i\kern-.025em b}\kern-.08em
    T\kern-.1667em\lower.7ex\hbox{E}\kern-.125emX}}
\begin{document}
\bstctlcite{IEEEexample:BSTcontrol}

\title{Parallelization of Monte Carlo Tree Search in Continuous Domains\\
\thanks{We wish to thank the German Research Foundation (DFG) for funding the project Cooperatively Interacting Automobiles (CoInCar) within which the research leading to this contribution was conducted. The information as well as views presented in this publication are solely the ones expressed by the authors.}
}

\author{\IEEEauthorblockN{1\textsuperscript{st} Karl Kurzer}
\IEEEauthorblockA{\textit{Karlsruhe Institute of Technology}\\
Karlsruhe, Germany \\
kurzer@kit.edu}
\and
\IEEEauthorblockN{2\textsuperscript{nd} Christoph H\"ortnagl}
\IEEEauthorblockA{\textit{Karlsruhe Institute of Technology}\\
Karlsruhe, Germany \\
uyedd@student.kit.edu}
\and
\IEEEauthorblockN{3\textsuperscript{rd} J. Marius Z\"ollner}
\IEEEauthorblockA{\textit{Karlsruhe Institute of Technology}\\
Karlsruhe, Germany \\
zoellner@kit.edu}
}

\maketitle

\begin{abstract}
Monte Carlo Tree Search (MCTS) has proven to be capable of solving challenging tasks in domains such as Go, chess and Atari.
Previous research has developed parallel versions of MCTS, exploiting today's multiprocessing architectures.
These studies focused on versions of MCTS for the discrete case.
Our work builds upon existing parallelization strategies and extends them to continuous domains.
In particular, leaf parallelization and root parallelization are studied and two final selection strategies that are required to handle continuous states in root parallelization are proposed.
The evaluation of the resulting parallelized continuous MCTS is conducted using a challenging cooperative multi-agent system trajectory planning task in the domain of automated vehicles.
\end{abstract}


\section{Introduction}
Monte Carlo Tree Search (MCTS), based on the UCT (Upper Confidence Bound for Trees) selection policy \cite{Kocsis2006}, has demonstrated its strength on numerous occasions facing problems with large branching factors.
AlphaGo, AlphaZero and their successor MuZero are among the most prominent examples, reaching super-human performance in the games of Go, chess, and Atari \cite{Silver2016,Silver2017,Schrittwieser2019}.

Prior work on parallelization has given insights with regard to benefits of using today's multi-threaded processors to speed up MCTS \cite{Cazenave2007,Chaslot2008a,Soejima2010,Enzenberger2010,Bourki2011,Rocki2011}.
However, due to its nature MCTS and the parallelization of it have been primarily studied for discrete state and action spaces.
Our work extends prior research on MCTS in continuous domains \cite{Couteoux2008,Rolet2009,Yee2016,Kurzer2018a,Moerland2018} to the parallel case and develops novel final selection strategies for root parallelization that are required to address these domains.

\section{Related Work}

\subsection{Monte Carlo Tree Search}
An exhaustive tree search can find the optimal trajectory through any Markov decision process (MDP) with a finite set of states and actions.
However, as the action space grows larger, so does the tree and it becomes prohibitively expensive to search for an optimal trajectory.
Tree Search combined with Monte Carlo sampling addresses this issue, by approximating the optimal solution asymptotically through sampling.
Monte Carlo Tree Search explores different trajectories (i.e. tuples of states $\statet$ and actions $\actiont$) through the MDP, with the goal of discovering the trajectory that maximizes the return $\return$ from the root state.

The return $\return$ of a trajectory $\trajectory$ equals its accumulated discounted reward $\rewardt$ at time step $t$, taking action $\actiont$ in state $\statet$ \cite{Sutton2018}, see \eqref{eq:Return}.

\begin{equation}\label{eq:Return}
R(\trajectory) = \sum_{(\statet, \actiont) \in \trajectory}{r_t(\statet,\actiont)}
\end{equation}

The mean of the returns over all trajectories $\trajectoryspace$ sampled from policy $\policy(\action|\state)$, starting in state $\state$ and taking action $\action$, is the Monte Carlo estimate of the action value $Q_\policy(\state, \action)$ \cite{Sutton2018}, see \eqref{eq:MonteCarloEstimate}.

\begin{equation}\label{eq:MonteCarloEstimate}
Q_\policy(s,a) = \frac{1}{|\trajectoryspace|}\sum_{\trajectory \in \trajectoryspace \sim \policy}{R(\trajectory)}
\end{equation}

Given an initial root state of the MDP, MCTS approximates the action value in four sequential steps for each iteration until a terminal condition is met (e.g. until a time budget or computational budget is exceeded).
Since MCTS is an anytime algorithm \cite{Browne2012}, it returns an estimate after the first iteration.

\subsubsection{Selection}
The UCT value \cite{Kocsis2006} for all explored actions from the current state is calculated during the selection phase, 
see \eqref{eq:UCT}, and the state action tuple with the maximum UCT value is selected.
This process repeats until a state is selected that has not been fully explored (i.e. not all available actions in the state have been expanded), see Fig. \ref{fig:SelectionExpansion}.

The first term in \eqref{eq:UCT} fosters exploitation of previously explored actions with high action values.
The second term guarantees that all actions for a given state are being expanded at least once,
with $\visitcount(\state)$ being the visit count for state $\state$ and $\visitcount(\state,\action)$ the number of times action $\action$ has been chosen in that state.
To balance the exploration-exploitation trade-off, a constant factor $c$ is used.

\begin{equation}\label{eq:UCT}
UCT(s,a) = Q_\policy(\state,\action)+c\sqrt{\frac{2\log{\visitcount(\state)}}{\visitcount(\state,\action)}}
\end{equation}

\subsubsection{Expansion}
\begin{figure}
	\centering
	\def\svgwidth{0.8\columnwidth}
\begingroup%
  \makeatletter%
  \providecommand\color[2][]{%
    \errmessage{(Inkscape) Color is used for the text in Inkscape, but the package 'color.sty' is not loaded}%
    \renewcommand\color[2][]{}%
  }%
  \providecommand\transparent[1]{%
    \errmessage{(Inkscape) Transparency is used (non-zero) for the text in Inkscape, but the package 'transparent.sty' is not loaded}%
    \renewcommand\transparent[1]{}%
  }%
  \providecommand\rotatebox[2]{#2}%
  \newcommand*\fsize{\dimexpr\f@size pt\relax}%
  \newcommand*\lineheight[1]{\fontsize{\fsize}{#1\fsize}\selectfont}%
  \ifx\svgwidth\undefined%
    \setlength{\unitlength}{476.86001515bp}%
    \ifx\svgscale\undefined%
      \relax%
    \else%
      \setlength{\unitlength}{\unitlength * \real{\svgscale}}%
    \fi%
  \else%
    \setlength{\unitlength}{\svgwidth}%
  \fi%
  \global\let\svgwidth\undefined%
  \global\let\svgscale\undefined%
  \makeatother%
  \begin{picture}(1,0.51374846)%
    \lineheight{1}%
    \setlength\tabcolsep{0pt}%
    \put(0,0){\includegraphics[width=\unitlength,page=1]{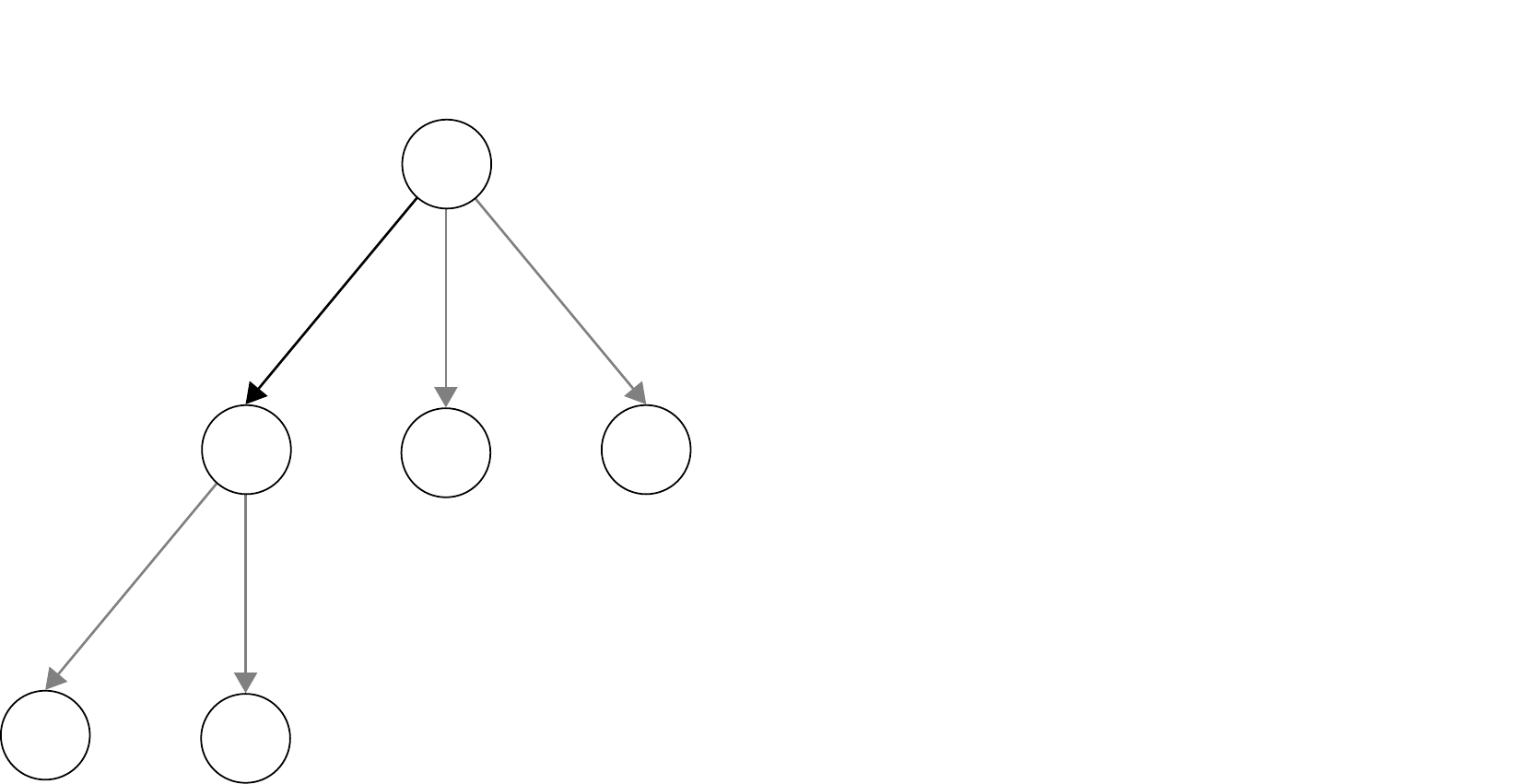}}%
    \put(0.15443098,0.48375183){\color[rgb]{0,0,0}\makebox(0,0)[lt]{\lineheight{1.25}\smash{\begin{tabular}[t]{l}Selection\end{tabular}}}}%
    \put(0,0){\includegraphics[width=\unitlength,page=2]{mcts_selection_expansion.pdf}}%
    \put(0.69216073,0.48984945){\color[rgb]{0,0,0}\makebox(0,0)[lt]{\lineheight{1.25}\smash{\begin{tabular}[t]{l}Expansion\end{tabular}}}}%
  \end{picture}%
\endgroup%

	\caption{Selection and Expansion in MCTS }
	\label{fig:SelectionExpansion}
\end{figure}

Once the selection policy encounters a state with untried actions left, it expands that state by randomly sampling an action from a uniform distribution over the action space, see \eqref{eq:UniformRandom}, and executing the action reaching a successor state, see Fig. \ref{fig:SelectionExpansion}.

\begin{equation}\label{eq:UniformRandom}
a \sim U[\min(\actionspace), \max(\actionspace)]
\end{equation}

\subsubsection{Simulation}
After the expansion of an action completes, a simulation of subsequent random actions is conducted
until a terminal condition is met (i.e. the planning horizon is reached or an action is sampled resulting in a terminal state) evaluating the action value of the previous expansion, see Fig. \ref{fig:SimulationBackpropagation}.

\subsubsection{Backpropagation}
\begin{figure}
	\centering
	\def\svgwidth{0.8\columnwidth}
\begingroup%
  \makeatletter%
  \providecommand\color[2][]{%
    \errmessage{(Inkscape) Color is used for the text in Inkscape, but the package 'color.sty' is not loaded}%
    \renewcommand\color[2][]{}%
  }%
  \providecommand\transparent[1]{%
    \errmessage{(Inkscape) Transparency is used (non-zero) for the text in Inkscape, but the package 'transparent.sty' is not loaded}%
    \renewcommand\transparent[1]{}%
  }%
  \providecommand\rotatebox[2]{#2}%
  \newcommand*\fsize{\dimexpr\f@size pt\relax}%
  \newcommand*\lineheight[1]{\fontsize{\fsize}{#1\fsize}\selectfont}%
  \ifx\svgwidth\undefined%
    \setlength{\unitlength}{476.85992864bp}%
    \ifx\svgscale\undefined%
      \relax%
    \else%
      \setlength{\unitlength}{\unitlength * \real{\svgscale}}%
    \fi%
  \else%
    \setlength{\unitlength}{\svgwidth}%
  \fi%
  \global\let\svgwidth\undefined%
  \global\let\svgscale\undefined%
  \makeatother%
  \begin{picture}(1,0.79587616)%
    \lineheight{1}%
    \setlength\tabcolsep{0pt}%
    \put(0,0){\includegraphics[width=\unitlength,page=1]{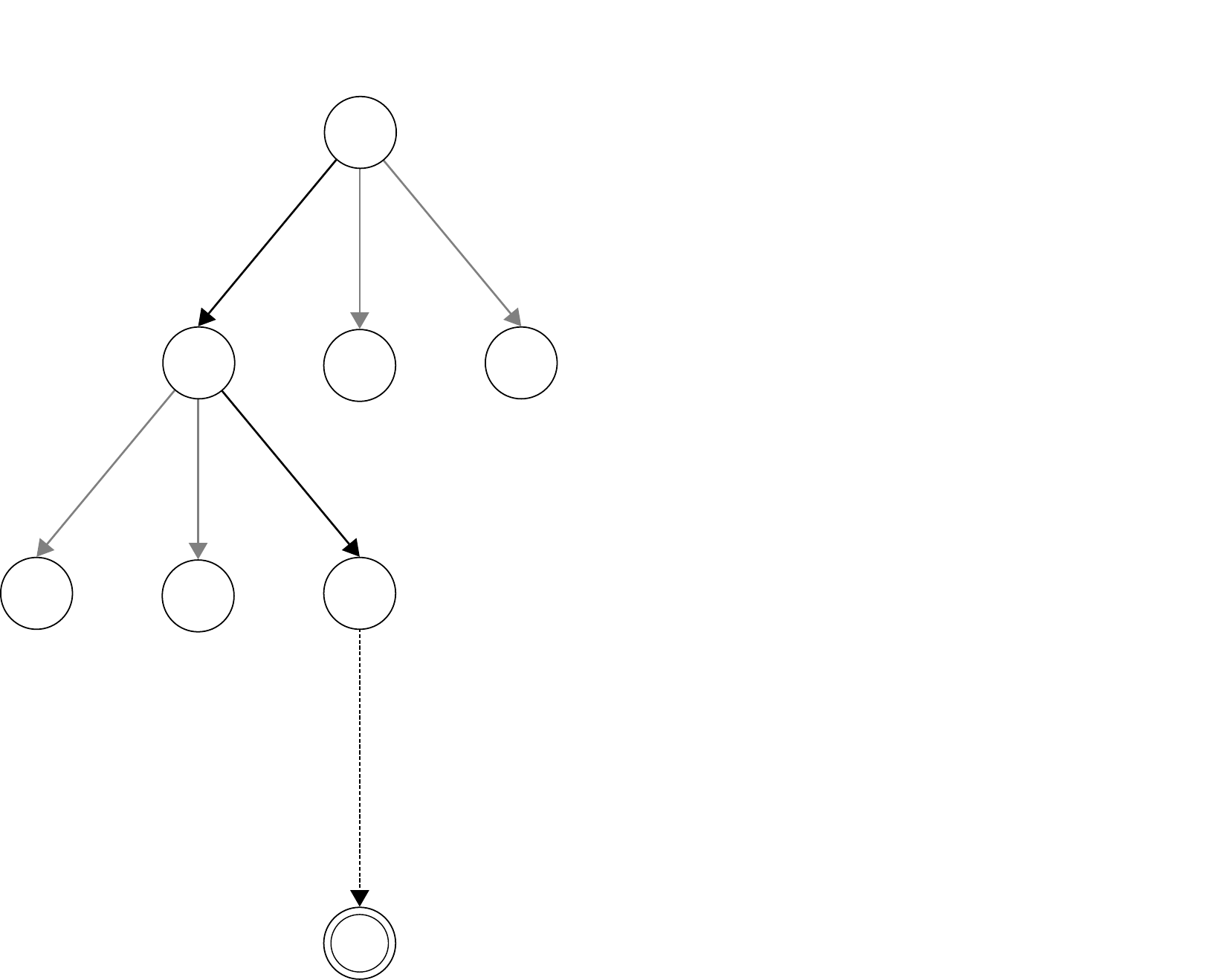}}%
    \put(0.1431266,0.76587952){\color[rgb]{0,0,0}\makebox(0,0)[lt]{\lineheight{1.25}\smash{\begin{tabular}[t]{l}Simulation\end{tabular}}}}%
    \put(0,0){\includegraphics[width=\unitlength,page=2]{mcts_simulation_backpropagation.pdf}}%
    \put(0.639847,0.77197714){\color[rgb]{0,0,0}\makebox(0,0)[lt]{\lineheight{1.25}\smash{\begin{tabular}[t]{l}Backpropagation\end{tabular}}}}%
  \end{picture}%
\endgroup%

	\caption{Simulation and Backpropagation in MCTS }
	\label{fig:SimulationBackpropagation}
\end{figure}

Lastly, the return $\return$ of the trajectory generated by the iteration is backpropagated to all states along the trajectory, see Fig. \ref{fig:SimulationBackpropagation},
and the action values and visit counts for all actions of the trajectory are updated, see \eqref{eq:UpdateVisit} and \eqref{eq:UpdateAction}, respectively.

\begin{equation}\label{eq:UpdateVisit}
\visitcount(\state,\action) = \visitcount(\state,\action) + 1
\end{equation}

\begin{equation}\label{eq:UpdateAction}
Q_\policy(\state,\action) = Q_\policy(\state,\action) + \frac{1}{\visitcount(\state,\action)}(R(\state,\action) - Q_\policy(\state,\action) )
\end{equation}

\subsection{Parallelization}
The parallelization of MCTS is a well researched topic.
Most of this research focuses on the game of Go \cite{Cazenave2007,Chaslot2008a,Soejima2010,Enzenberger2010,Bourki2011,Rocki2011}.
The baseline version of MCTS can be parallelized in multiple ways\textsl{}.
Three different types of MCTS parallelizations are commonly referred to, namely leaf parallelization, root parallelization, and tree parallelization \cite{Browne2012,Chaslot2008a}.

\subsubsection{Leaf Parallelization}
The simplest form of parallelization is leaf parallelization.
In this case all policies behave identical except for the simulation policy.
Once a node gets expanded each thread creates a copy of the expanded node and runs a simulation on the copy until it terminates (i.e. reaches a terminal state or the planning horizon).
The results of all simulations are combined and backpropagated along the traversed branch.
One drawback of this method is the synchronization of threads before the backpropagation,
as this implies that all threads must wait for the slowest to finish.
By design, leaf parallelization fosters exploitation.

\begin{figure}
	\fns
	\centering
	\def\svgwidth{0.25\columnwidth}
	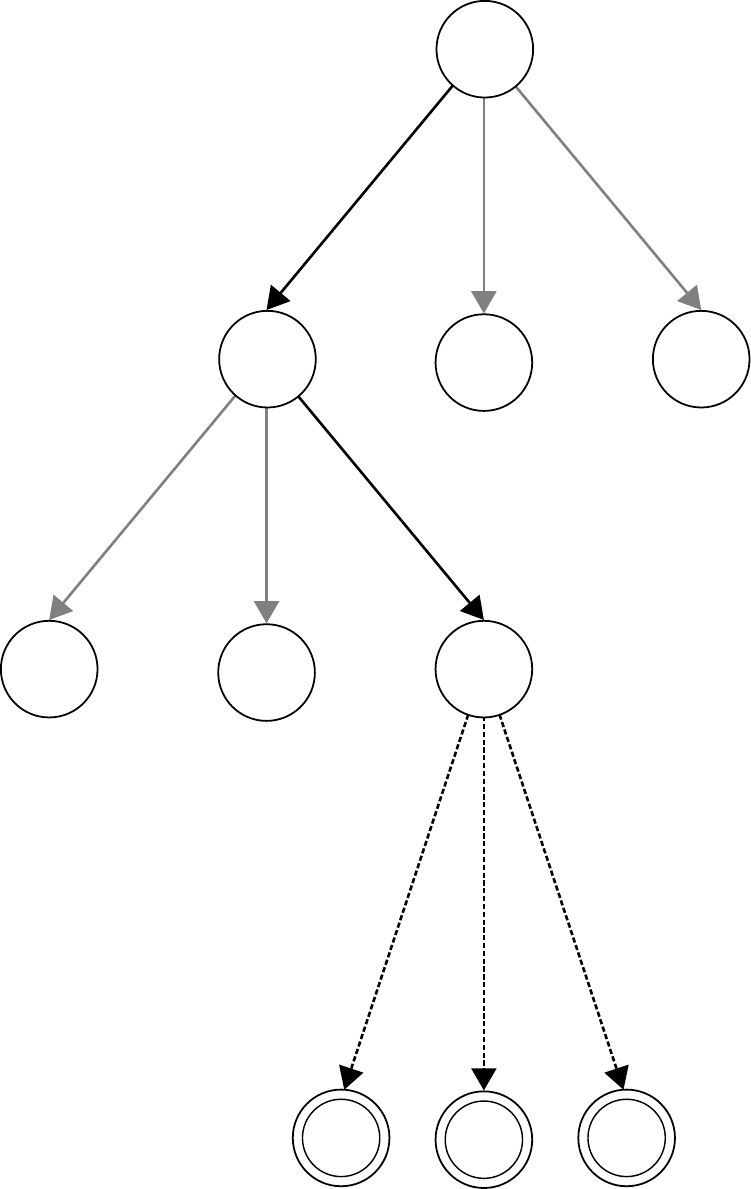
	\caption{Leaf Parallelization of MCTS}
	\label{fig:MCTSLeaf}
\end{figure}

\subsubsection{Root Parallelization}
Contrary to leaf parallelization root parallelization fosters exploration, even though not in a principled way. 
During root parallelization each thread creates a copy of the root node and builds up its own tree.
Once the allotted time or computational budget is reached, the resulting trees are merged.
Different methods on how to merge the resulting trees have been proposed.
Some studies use a voting mechanism, where each tree votes for the best action it has explored \cite{Soejima2010} 
and the action with the majority of the votes gets selected.

\begin{figure}
	\fns
	\centering
	\def\svgwidth{\columnwidth}
\begingroup%
  \makeatletter%
  \providecommand\color[2][]{%
    \errmessage{(Inkscape) Color is used for the text in Inkscape, but the package 'color.sty' is not loaded}%
    \renewcommand\color[2][]{}%
  }%
  \providecommand\transparent[1]{%
    \errmessage{(Inkscape) Transparency is used (non-zero) for the text in Inkscape, but the package 'transparent.sty' is not loaded}%
    \renewcommand\transparent[1]{}%
  }%
  \providecommand\rotatebox[2]{#2}%
  \newcommand*\fsize{\dimexpr\f@size pt\relax}%
  \newcommand*\lineheight[1]{\fontsize{\fsize}{#1\fsize}\selectfont}%
  \ifx\svgwidth\undefined%
    \setlength{\unitlength}{904.92957408bp}%
    \ifx\svgscale\undefined%
      \relax%
    \else%
      \setlength{\unitlength}{\unitlength * \real{\svgscale}}%
    \fi%
  \else%
    \setlength{\unitlength}{\svgwidth}%
  \fi%
  \global\let\svgwidth\undefined%
  \global\let\svgscale\undefined%
  \makeatother%
  \begin{picture}(1,0.2297082)%
    \lineheight{1}%
    \setlength\tabcolsep{0pt}%
    \put(0,0){\includegraphics[width=\unitlength,page=1]{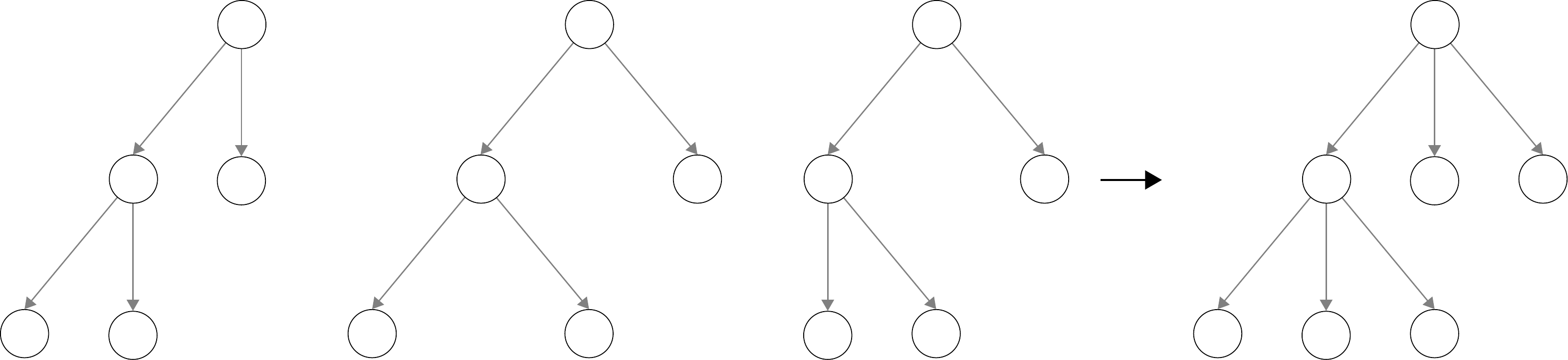}}%
    \put(0.47723112,0.1079259){\color[rgb]{0,0,0}\makebox(0,0)[lt]{\lineheight{1.25}\smash{\begin{tabular}[t]{l}+\end{tabular}}}}%
    \put(0.18659756,0.1079259){\color[rgb]{0,0,0}\makebox(0,0)[lt]{\lineheight{1.25}\smash{\begin{tabular}[t]{l}+\end{tabular}}}}%
  \end{picture}%
\endgroup%

	\caption{Root Parallelization of MCTS}
	\label{fig:MCTSRoot}
\end{figure}

\subsubsection{Tree Parallelization}
Another variant is tree parallelization, which can be seen as a mixture of the previous methods, as the search tree is shared for all phases of the MCTS.
However, as multiple threads build up the same tree it is essential to coordinate the threads properly ensuring efficiency as well as effectiveness of the parallel architecture.
Thus, lock-free approaches have been developed to implement tree parallelization efficiently \cite{Enzenberger2010,Mirsoleimani2018}.
In addition, measures such as virtual loss \cite{Chaslot2008a} improve the effectiveness of tree parallelization, by increasing the likelihood that multiple threads traverse different paths.
While virtual loss has been an adopted concept \cite{Silver2016}, more recent studies suggest that the increase in effectiveness is merely a trade-off with time efficiency \cite{Mirsoleimani2017}.

Even though the proposed parallelization strategies reach performance gains, the inherent dependence on previous simulation results in UCT remains a key issue that hinders scalability. In order to reduce the dependence on previous iterations, the visit count for incomplete expansions/simulations can be incorporated in the UCT formula.
This avoids multiple threads working on identical unexplored actions and increases the information content of UCT prior to the result of the simulation \cite{Liu2020}.

As can be seen, extensive research on the parallelization of MCTS exists for discrete domains.
However, to the best of our knowledge no research on the parallelization of MCTS with continuous action and state spaces has been performed.
Hence, we propose respective extensions which are required by continuous domains.

\section{Approach}

\subsection{Extensions to Continuous Action Spaces}
The application of the UCT in MCTS demands that every possible action in a given node must be explored at least once \cite{Kocsis2006}.
When actions are drawn from a continuous distribution, MCTS would be stuck indefinitely exploring a single node.
Thus, standard MCTS is not directly applicable to continuous domains.

\subsubsection{Progressive Widening}
To address the before mentioned issue during the expansion phase in continuous action domains, \textit{Progressive Unpruning} \cite{Chaslot2008b} or \textit{Progressive Widening} \cite{Coulom2007} have been proposed. Additional theoretical considerations have extended UCT for continuous domains \cite{Wang2009}.

These approaches gradually add actions during runtime to the action space of a  node.
Each node starts out with a sampled set of actions.
\textit{Progressive Widening} takes place once a node has sufficiently been explored
by sampling an additional action and adding it to the action space of the node.
The criteria is defined by a sub-linear function, see \eqref{eq: progressive widening criteria}.

\begin{equation} \label{eq: progressive widening criteria}
|\actionspace(\state)| = \lfloor c \cdot \visitcount(s)^{\alpha} \rfloor
\end{equation}
The cardinality of the action set for any given state is dependent on the visit count 
of the state itself.
The choice of the parameters $c$ and $\alpha \in[0,1]$ is adapted empirically.

While progressive widening circumvents the problem of continuous action spaces during expansion,
it does not solve the inefficiency in the update phase.

\subsubsection{Similarity Update}
Since the update phase aggregates the statistics from consecutive simulations to generate accurate
value estimates, it is critical to use as many simulations as possible for variance reduction.
In continuous action spaces it regularly occurs that actions are sampled in close vicinity to one another.
Actions that are close to each other are likely to generate similar trajectories and hence comparable returns.

\begin{figure}
	\centering
	\def\svgwidth{0.8\columnwidth}
	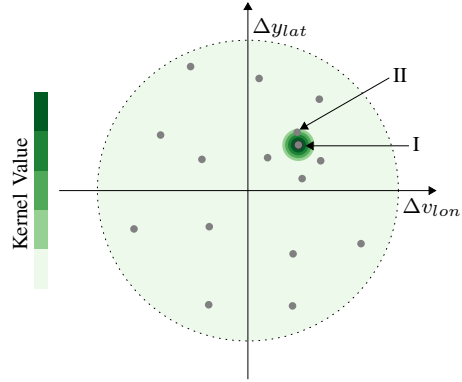
	\caption{Similarity update of action II given action I in the continuous space; In order to share information between actions that are similar, action II is updated with statistics gathered from action I weighted by the kernel value of action I and II.}
	\label{fig:SimilarityUpdate}
\end{figure}

An example of this is depicted in Fig.~\ref{fig:SimilarityUpdate}, where action I has been sampled in close vicinity to action II, thus the return of action I might hold valuable information for action II and could be integrated in its statistics, to reduce its variance.

When the tree is traversed in reverse order during backpropagation, the similarity between the current action and all previously explored actions of a state is determined.
The return and the visit count of the current action are weighted with the similarity to update comparable states. 

The similarity of two actions $\similarity(\action_i, \action_j) \in (0,1]$ in the action space is determined by a distance measure based on a radial basis function, see \eqref{eq:similarity update}.
Lower values for $\gamma \in \mathbb{R}^+$, increase the influence of other actions, higher values decrease it.

\begin{equation}\label{eq:similarity update}
\similarity(\action_i, \action_j) =  \exp{\left ( -\gamma \left \Vert \action_i - \action_j \right \Vert^2 \right )}
\end{equation}

\subsection{Leaf Parallelization}
As mentioned, the simplest form of parallelization can be conducted on the leaves of the tree.
The node to be simulated is copied once for each thread and the simulation is run until all threads reach the terminal condition.
Finally, the result of the simulations is aggregated.

\begin{equation}\label{eq:r_sim}
\reward_\text{sim} = \sum_{t}^{T}{\discountfactor^t\reward(\state,\action)}
\end{equation}

\subsubsection{Mean}\label{sec:leafMean}
The most obvious aggregation is the mean simulation reward, which equals the cumulated sum of rewards over all time steps $t$ over all threads $\thread \in \threadspace$, see \eqref{eq:leaf avg}.

\begin{equation}\label{eq:leaf avg}
\overline{\reward}_\text{sim} = \frac{1}{|\threadspace|} \sum_{\thread}^{\threadspace}\reward_\text{sim}(\thread)
\end{equation}

One issue arising with the mean is that environments requiring precise action selection become harder to solve \cite{Soemers2016}.
As the majority of actions from a given state leads to an undesirable terminal state
and only a fraction result in the desired state, averaging over all simulations will lead to a pessimistic evaluation of that node.

\subsubsection{Maximum}\label{sec:leafMax}
To overcome the problem, the maximum of the simulation reward can be used, however,
this cannot be generalized to adversarial multiplayer environments, as it might lead to traps \cite{Soemers2016}.
The maximum simulation reward is calculated by choosing the value from the thread $\thread$ that generated the maximum cumulated sum of rewards over all time steps $t$ during simulation \eqref{eq:leaf max}.

\begin{equation}\label{eq:leaf max}
\reward^*_\text{sim} = \max_\thread\reward_\text{sim}(\thread)
\end{equation}

\subsection{Root Parallelization}
Root parallelization grows multiple trees $\treespace$ originating from the same root.
Once the MCTS terminates, the statistics of all resulting trees need to be merged.
Merging the information for final action selection in the continuous domain requires different strategies, as actions that were explored in one tree are unlikely to be found in another.
In the following two novel methods for aggregating the statistics are presented.

\subsubsection{Similarity Merge}
Using the similarity update depicted in Fig. \ref{fig:SimilarityUpdate} multiple trees can be merged into a single tree, from which a final action can be chosen.

The process is described in Algorithm \ref{alg:SimilarityMerge}. First, all actions of all trees are added to the final tree, line \ref{sm:t_final}. Second, a pairwise similarity update is conducted between all actions of the final tree and the simulated trees.
For this, the similarity visit count is determined, line \ref{sm:sm_visit}, and the similarity action value is calculated based on the weighted visit count and similarity, line \ref{sm:sm_value}. Last, the action with the maximum action value from the final tree is returned.

\begin{algorithm}
	\caption{Similarity Merge}
	\label{alg:SimilarityMerge}
	\begin{algorithmic}[1]
		\algsetup{linenodelimiter=}
		\REQUIRE $\simmat$
		\FOR{$\tree \in \treespace$}
			\STATE $\actionspace_{\tree_\text{final}} \leftarrow \actionspace_{\tree_\text{final}} + \actionspace_\tree$ \label{sm:t_final}
		\ENDFOR
		\FOR{$\action_i \in \actionspace_{\tree_\text{final}}$}
		\FOR{$\action_j \neq \action_i \in \actionspace_{\tree_\text{final}}$}
		\STATE $\simmat_{ij} \leftarrow \exp{\left ( -\gamma \left \Vert \action_i - \action_j \right \Vert^2 \right )}$
			\STATE $\visitcount_\text{sim}(\state_0,\action_i) \leftarrow \visitcount(\state_0,\action_i) + \simmat_{ij}\visitcount(\state_0,\action_j)$ \label{sm:sm_visit}
			\STATE $Q_\text{sim}(\state_0, a_i) \leftarrow \linebreak \frac{1}{\visitcount_\text{sim}} (\visitcount(\state_0,\action_i)Q(\state_0, a_i) + \simmat_{ij}\visitcount(\state_0,\action_j)Q(\state_0, a_j))$ \label{sm:sm_value}
			\ENDFOR
			\ENDFOR
			\RETURN $\action_\text{final} \leftarrow \argmax_{\action \in \actionspace_{\tree_\text{final}}} Q_\text{sim}(\state_0, \action) $
	\end{algorithmic}
\end{algorithm}

\subsubsection{Similarity Vote}
Inspired by a voting scheme for root parallelization which does not merely choose the action with the overall highest number of visits or action value, but rather lets each tree vote for an action \cite{Soejima2010}, an extension for continuous domains was developed, see Algorithm \ref{alg:SimilarityVote}.
Analogous to \cite{Soejima2010} each tree submits its best action, line \ref{sv:submission}.
Then a similarity matrix $\simmat$ is calculated, that stores the similarity of a chosen action from one tree to all actions from all trees, line \ref{sv:simmat}.
Weighted by the action values of the submitted actions, line \ref{sv:vote}, the final action is the one that maximizes the similarity vote.

\begin{algorithm}
	\caption{Similarity Vote}
	\label{alg:SimilarityVote}
	\begin{algorithmic}[1]
	\algsetup{linenodelimiter=}
	\REQUIRE $\simmat, \votemat$
			\FOR{$\tree \in \treespace$}
				\STATE $\actionspace^* \leftarrow \actionspace^* \cup \argmax_{\action} \label{sv:submission} Q_\tree(\state_0,\action)$
			\ENDFOR
			\FOR{$\action_i \in \actionspace^*$}
				\FOR{$\action_j \in \actionspace^*$}
				\STATE $\simmat_{ij} \leftarrow \exp{\left ( -\gamma \left \Vert \action_i - \action_j \right \Vert^2 \right )}$\label{sv:simmat}
				\ENDFOR
			\STATE $\votemat_{i} \leftarrow Q(\state_0, a_i)$ \label{sv:vote}
			\ENDFOR
			\RETURN $\action_\text{final} \leftarrow a_i \in \actionspace^* | i = \argmax_i \simmat\votemat$
	\end{algorithmic}
\end{algorithm}

\section{Evaluation}
To evaluate the impact of the proposed parallelization techniques, a challenging multi-agent environment, i.e. cooperative trajectory planning for automated vehicles \cite{Kurzer2018a} is employed.
Using MCTS in conjunction with Decoupled-UCT, actions for each agent are evaluated in a decentralized manner, while respecting the interdependence of actions among traffic participants.

\begin{figure}
	\fns
	\begin{subfigure}{\columnwidth}
		\centering
		\def\svgwidth{\columnwidth}
		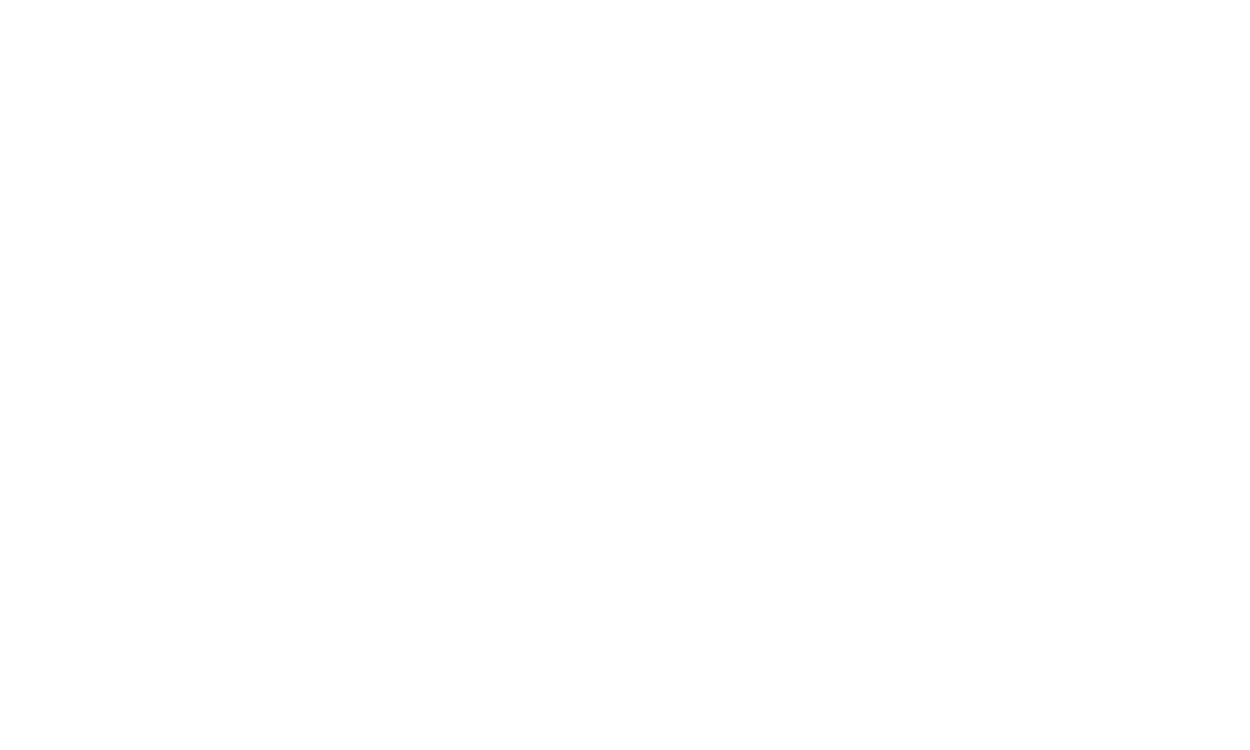
		\caption{Leaf Parallelization: Mean}
		\label{fig:MCTSLeafMean}
	\end{subfigure}
	\begin{subfigure}{\columnwidth}
		\centering
		\def\svgwidth{\columnwidth}
		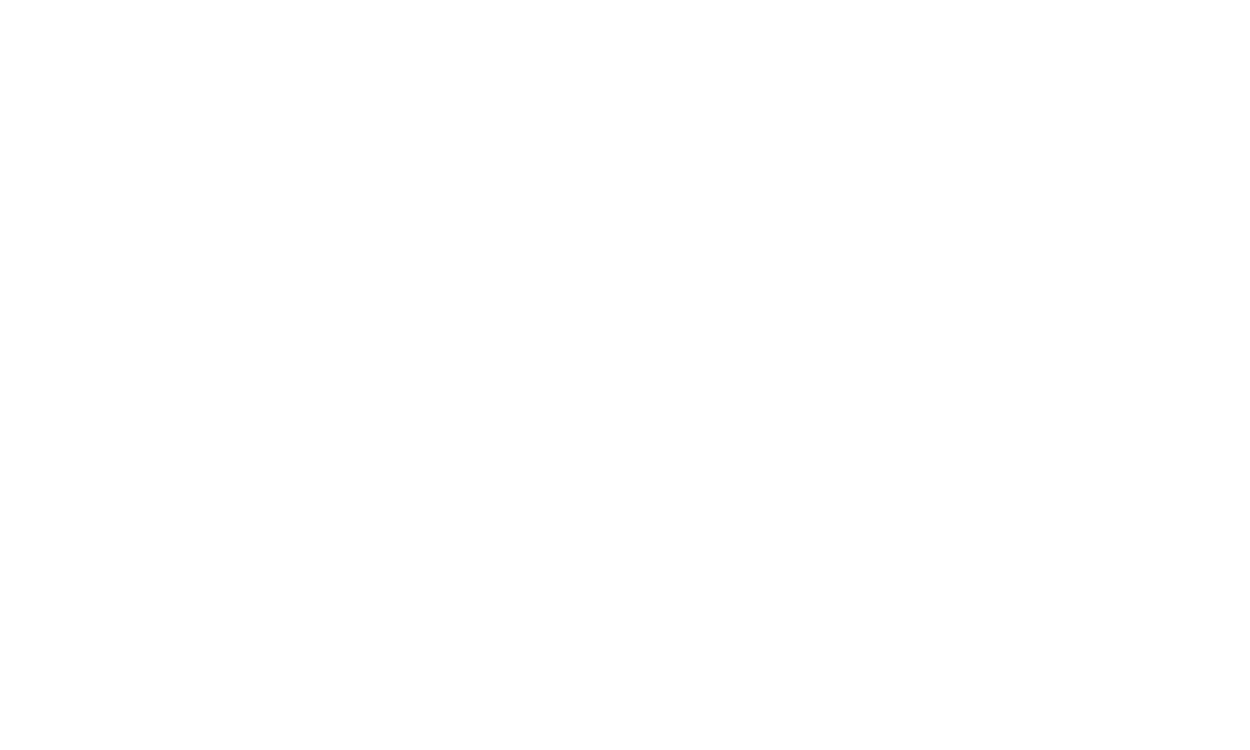
		\caption{Leaf Parallelization: Maximum}
		\label{fig:MCTSLeafMax}
	\end{subfigure}
	\begin{subfigure}{\columnwidth}
		\centering
		\def\svgwidth{\columnwidth}
		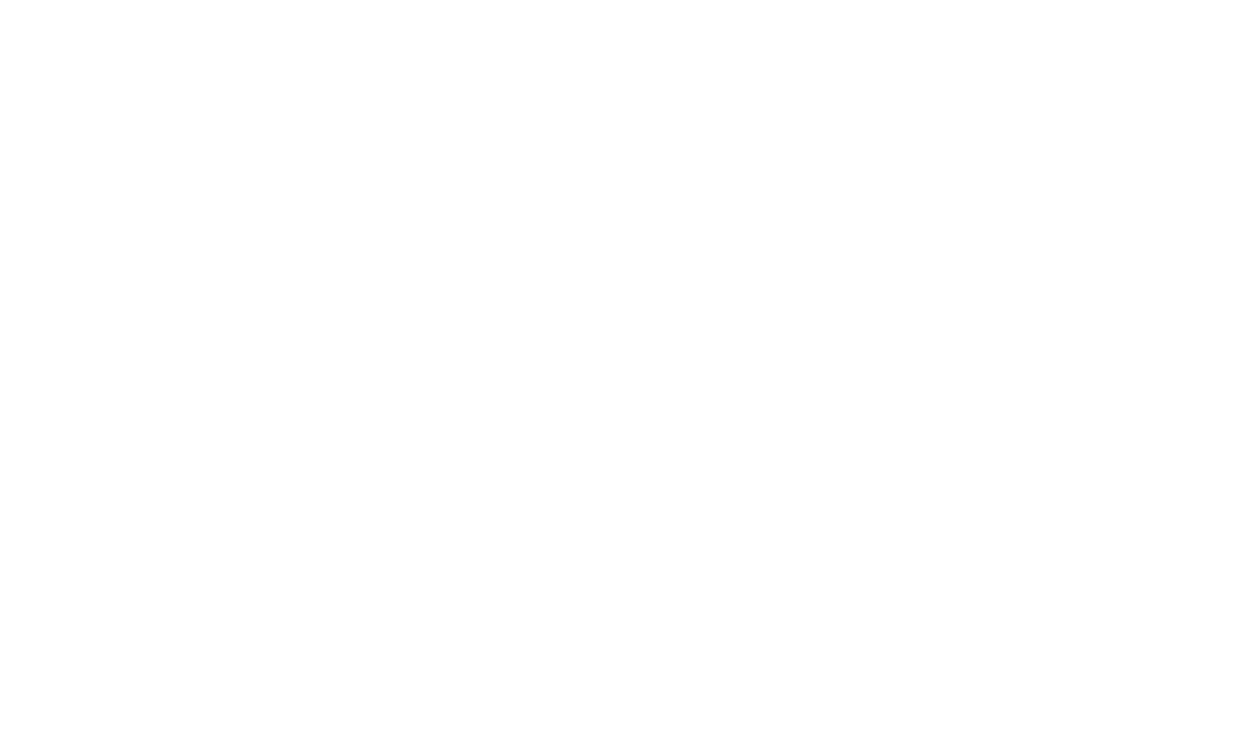
		\caption{Root Parallelization: Similarity Merge}
		\label{fig:MCTSRootMergeG5}
	\end{subfigure}
	\begin{subfigure}{\columnwidth}
		\centering
		\def\svgwidth{\columnwidth}
		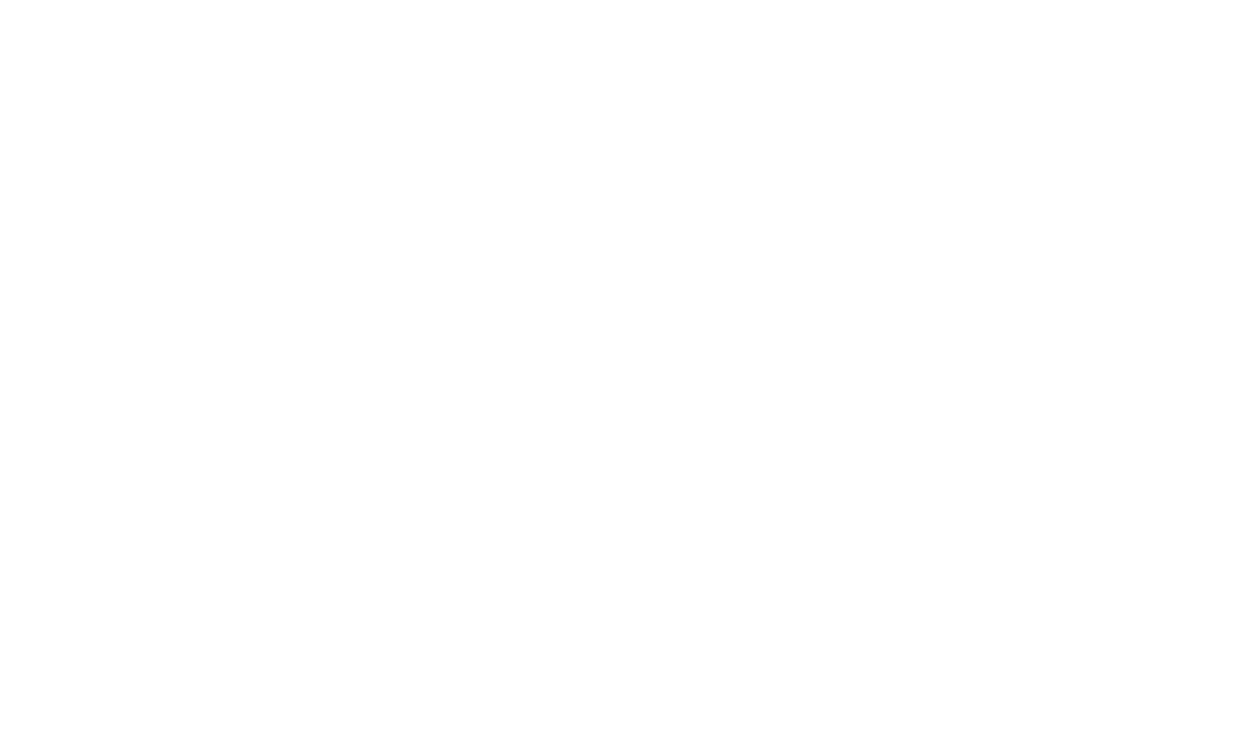
		\caption{Root Parallelization: Similarity Vote}
		\label{fig:MCTSRootVoteG1}
	\end{subfigure}
	\caption{Scalability of the different proposed parallelization strategies; The shaded region represents the deviation ($2\sigma$), of the mean success rate for the single-threaded baseline. The x-axis uses a logarithmic scale.}
	\label{fig:MCTSScalability}
\end{figure}

\begin{figure}
	\fns
	\begin{subfigure}{\columnwidth}
		\centering
		\def\svgwidth{\columnwidth}
		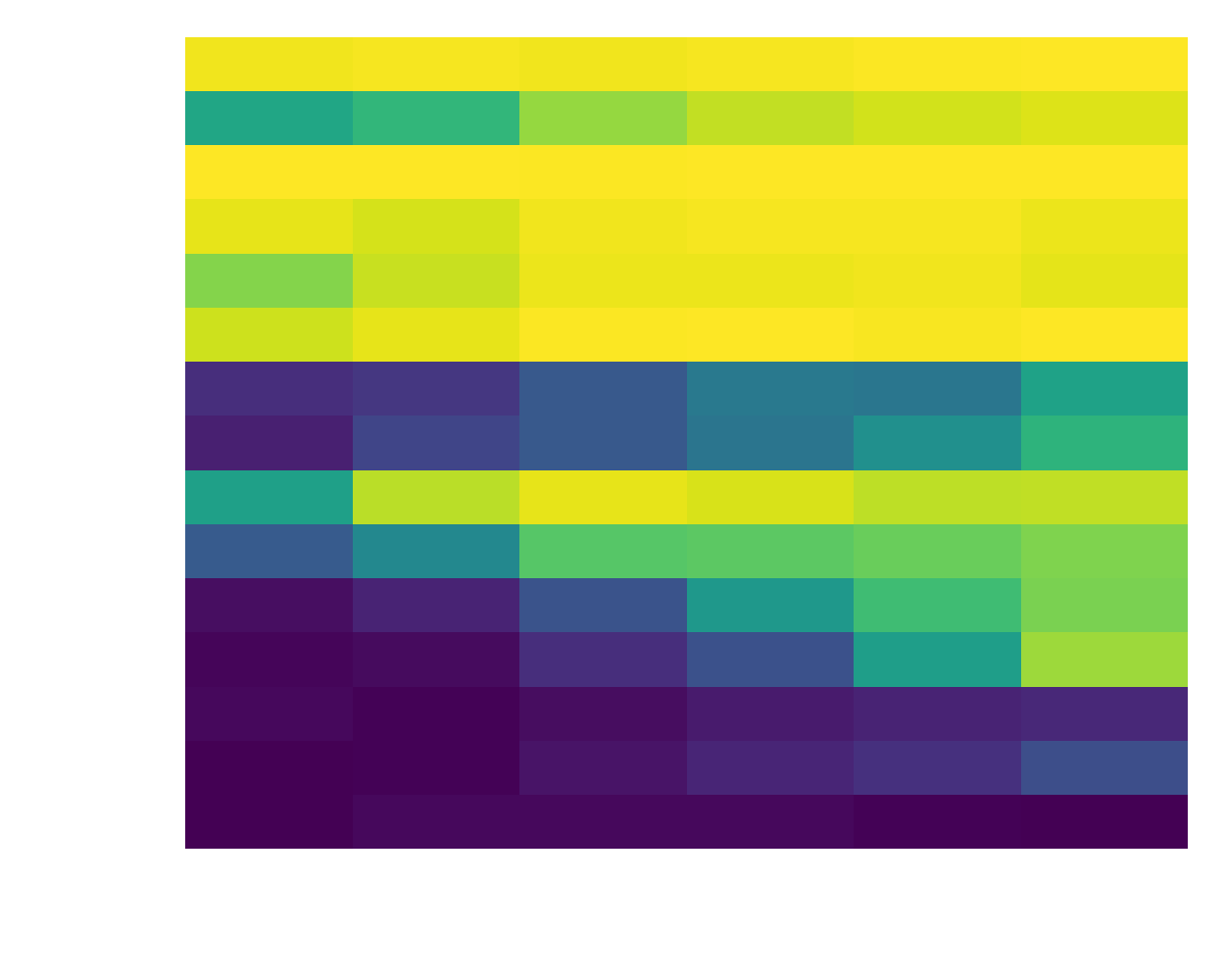
		\caption{1 Thread}
	\label{fig:MCTSSingleThreadedDetailed}
	\end{subfigure}
	\begin{subfigure}{\columnwidth}
		\centering
		\def\svgwidth{\columnwidth}
		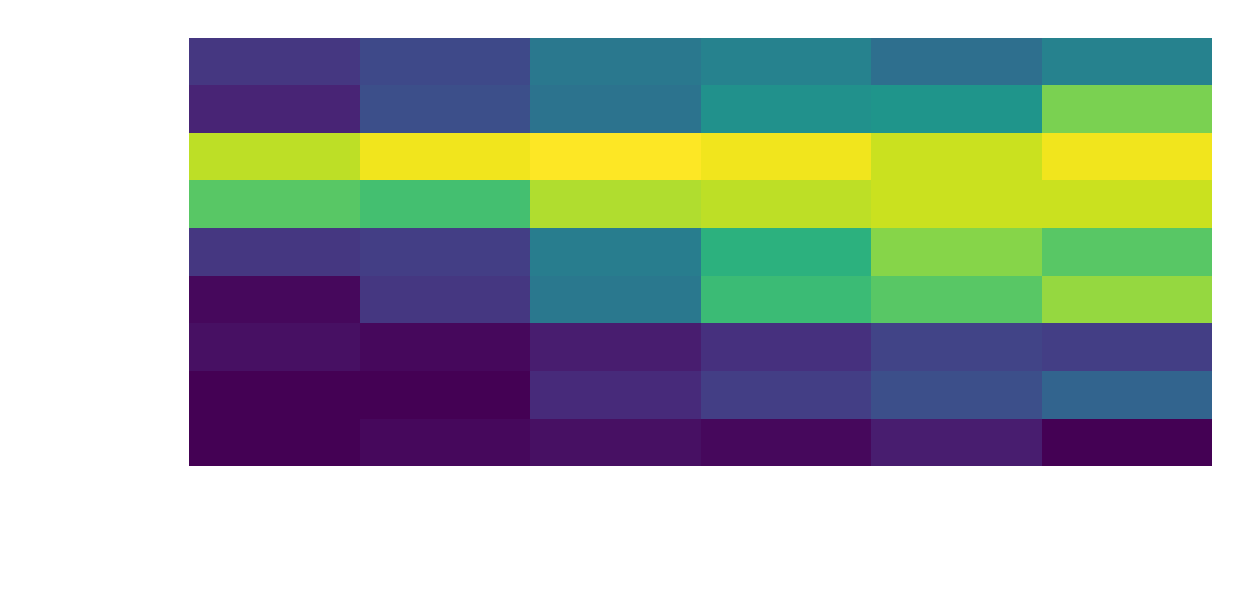
		\caption{8 Threads}
	\end{subfigure}
	\begin{subfigure}{\columnwidth}
		\centering
		\def\svgwidth{\columnwidth}
		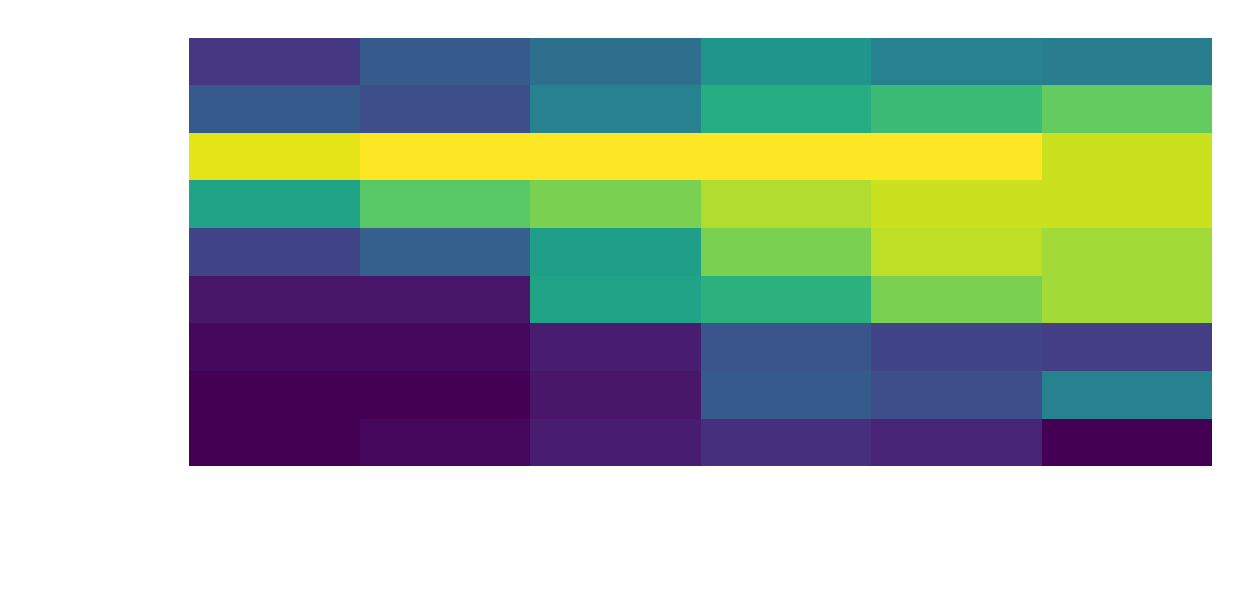
		\caption{24 Threads}
	\end{subfigure}
	\caption{Evaluation of the success rate (i.e. $1-\text{collision rate}$) for the root parallelization with similarity vote for 1, 8 and 24 threads}
	\label{fig:MCTSRootVoteG1Detailed}
\end{figure}

Each parallelization technique is evaluated on every scenario for different numbers of iterations and threads.
To generate statistically reliable results each setting is evaluated 50 times.
The results for the scalability analysis are summarized in Fig. \ref{fig:MCTSScalability}.
The key evaluation metric is the success rate for the scenarios (the inverse of the collision rate, i.e. whether or not a vehicle collided).
All parallelization strategies are compared to the performance of the average single-threaded baseline.
Unless stated otherwise, the evaluation is conducted using scenarios SC07--SC15.
This avoids skewing of the results as the single-threaded case already reaches a success rate of close to 100 percent for scenarios SC01--SC06, cf. Fig. \ref{fig:MCTSSingleThreadedDetailed}, allowing for no improvement through parallelization.
Videos of all scenarios combined with a brief description can be found online \footnote{\url{http://url.fzi.de/PC-MCTS-COG}}.

\begin{figure}
	\fns
		\centering
		\def\svgwidth{\columnwidth}
		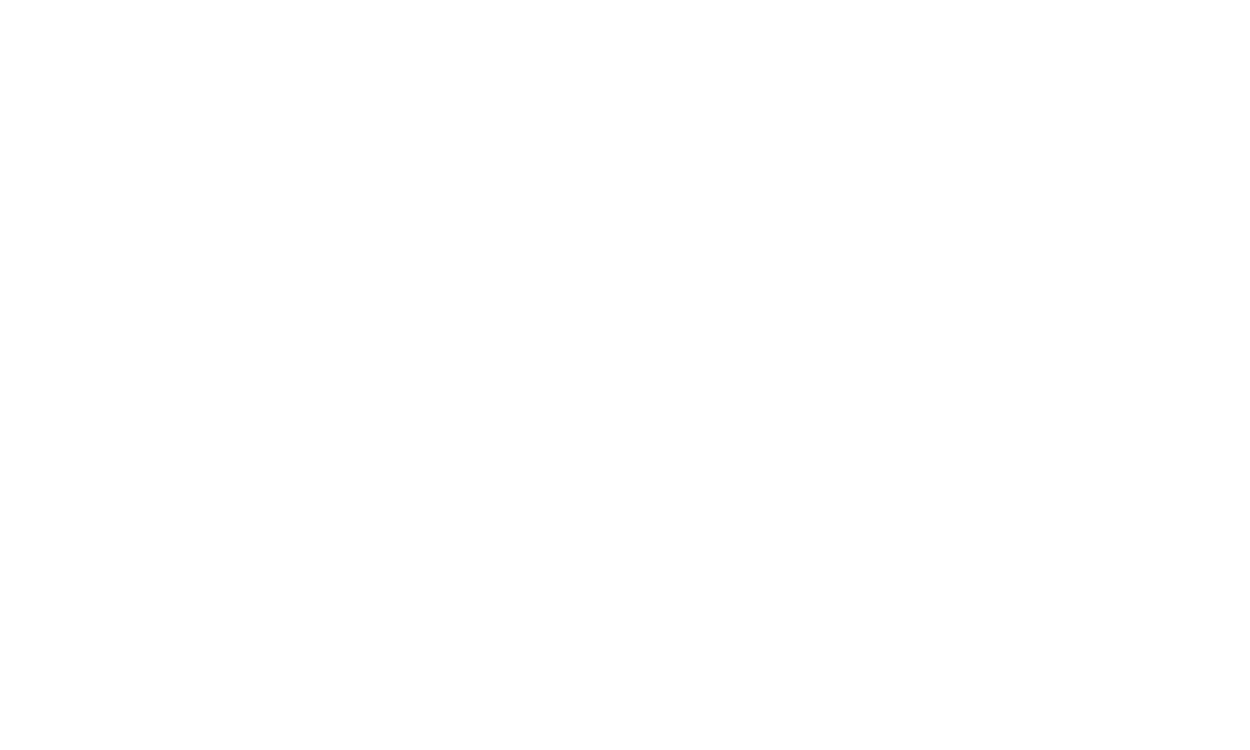
	\caption{Comparison of the different proposed parallelization strategies; The shaded region represents the deviation ($2\sigma$), of the mean success rate for the single-threaded baseline. The x-axis uses a logarithmic scale.}
	\label{fig:MCTSParallelComparison}
\end{figure}

\subsection{Leaf Parallelization}

Fig. \ref{fig:MCTSLeafMean} depicts the results of leaf parallelization with mean aggregation.
No improvement and no scaling of threads can be seen.
The slight gain in success rate over the $2\sigma$ region of the baseline at 2000 iterations is not consistent as it vanishes again for 4000 iterations.

While the results for the maximum aggregation depicted in Fig. \ref{fig:MCTSLeafMax} indicate that it does not scale well with the number of threads, they demonstrate that there is relatively consistent improvement for 8 threads, as the success rate stays outside the $2\sigma$ region of the baseline.
This implies that there is a benefit for multiple roll-outs when using their maximum.
However, this might be due to the nature of the cooperative environment as explained in section \ref{sec:leafMax}.

\subsection{Root Parallelization}
Similarity merge does neither perform well, nor does it scale, see Fig. \ref{fig:MCTSRootMergeG5}. Different parameter values for $\gamma$, led to comparable results.
 
Results for the similarity vote indicate a performance gain for lower iteration counts which seems to increase until 1000 iterations and tapers off afterwards, pictured in Fig. \ref{fig:MCTSRootVoteG1}.
A detailed evaluation for the similarity vote is shown in Fig. \ref{fig:MCTSRootVoteG1Detailed}.
For lower numbers of iterations the improvement is considerable.
At 100 iterations the success rate improves by approximately 100 percent, and keeps an improvement of roughly 45 percent until reaching 1000 iterations.

\section{Conclusion and Outlook}
The parallelization of MCTS in continuous domains creates novel challenges, which is especially true for root parallelization.
Different solutions were demonstrated and evaluated to address these challenges.
Standard leaf parallelization with mean selection, which has been shown to perform poorly in discrete domains seems to be less promising.
However, the modification to maximum selection performed considerably better, which was to be expected in a cooperative multi-agent environment.
Root parallelization with similarity merge yields no significant improvement.
In combination with similarity voting the performance of root parallelization improved considerably, especially for lower iteration counts.
Overall similarity voting performed best, cf. Fig. \ref{fig:MCTSParallelComparison}.

While the proposed solutions achieve higher success rates, they need to be optimized in order to be more time efficient, as the allocation of threads does incur a considerable overhead, especially when the number of threads increases.

In our future research we will look into additional techniques, promising a speed-up which scales better with the number of threads \cite{Liu2020}, as the standard techniques for parallelization do not address the trade of between exploration and exploitation well enough \cite{Auer2003,Liu2020,Mirsoleimani2017}.

\begin{figure}
	\fns
	\begin{subfigure}{\columnwidth}
		\centering
		\def\svgwidth{\columnwidth}
		\input{./03_graphics/single_threadedsingle_threaded.pdf_tex}
		\caption{1 Thread}
	\end{subfigure}
	\begin{subfigure}{\columnwidth}
		\centering
		\def\svgwidth{\columnwidth}
		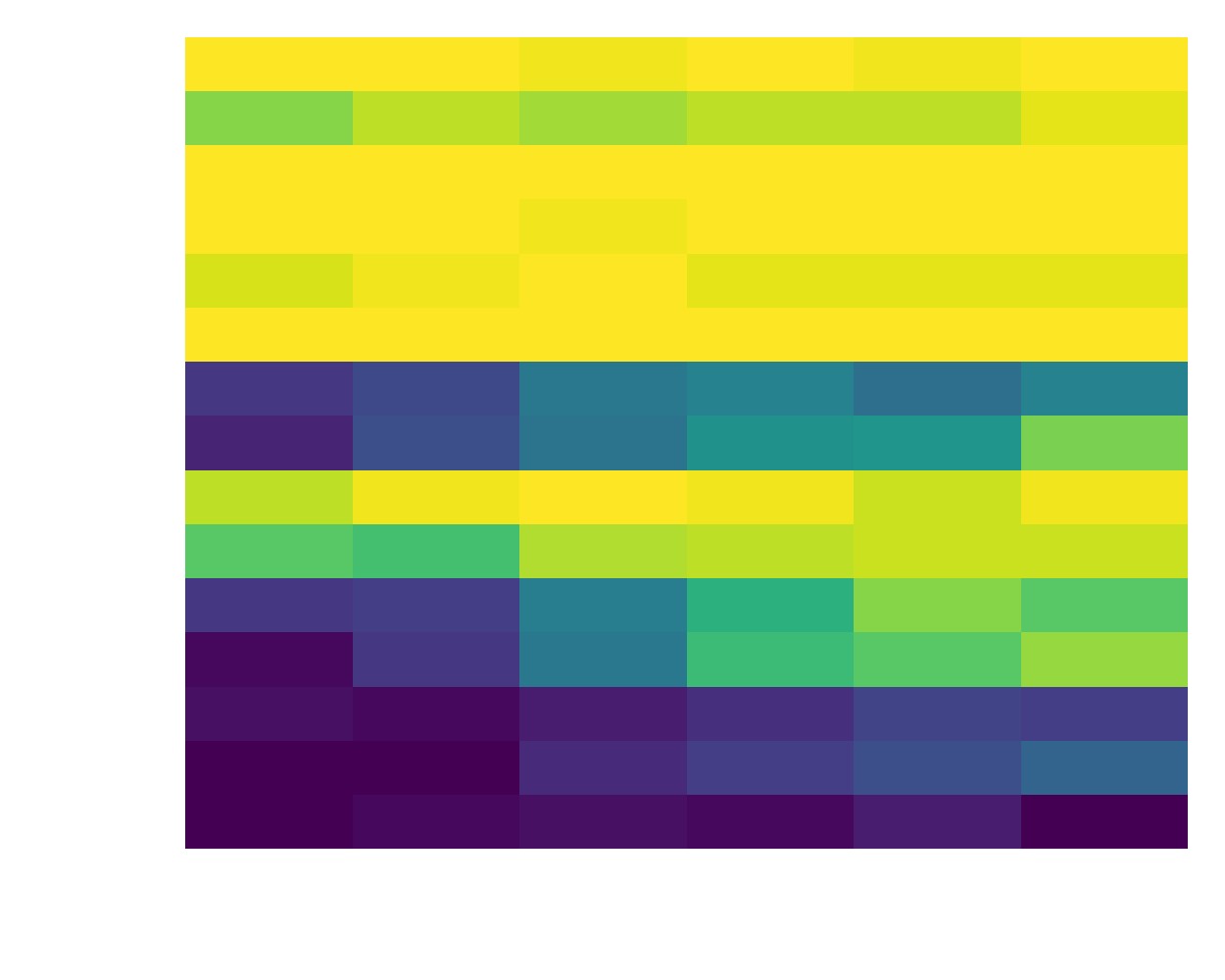
		\caption{8 Threads}
	\end{subfigure}
	\begin{subfigure}{\columnwidth}
		\centering
		\def\svgwidth{\columnwidth}
		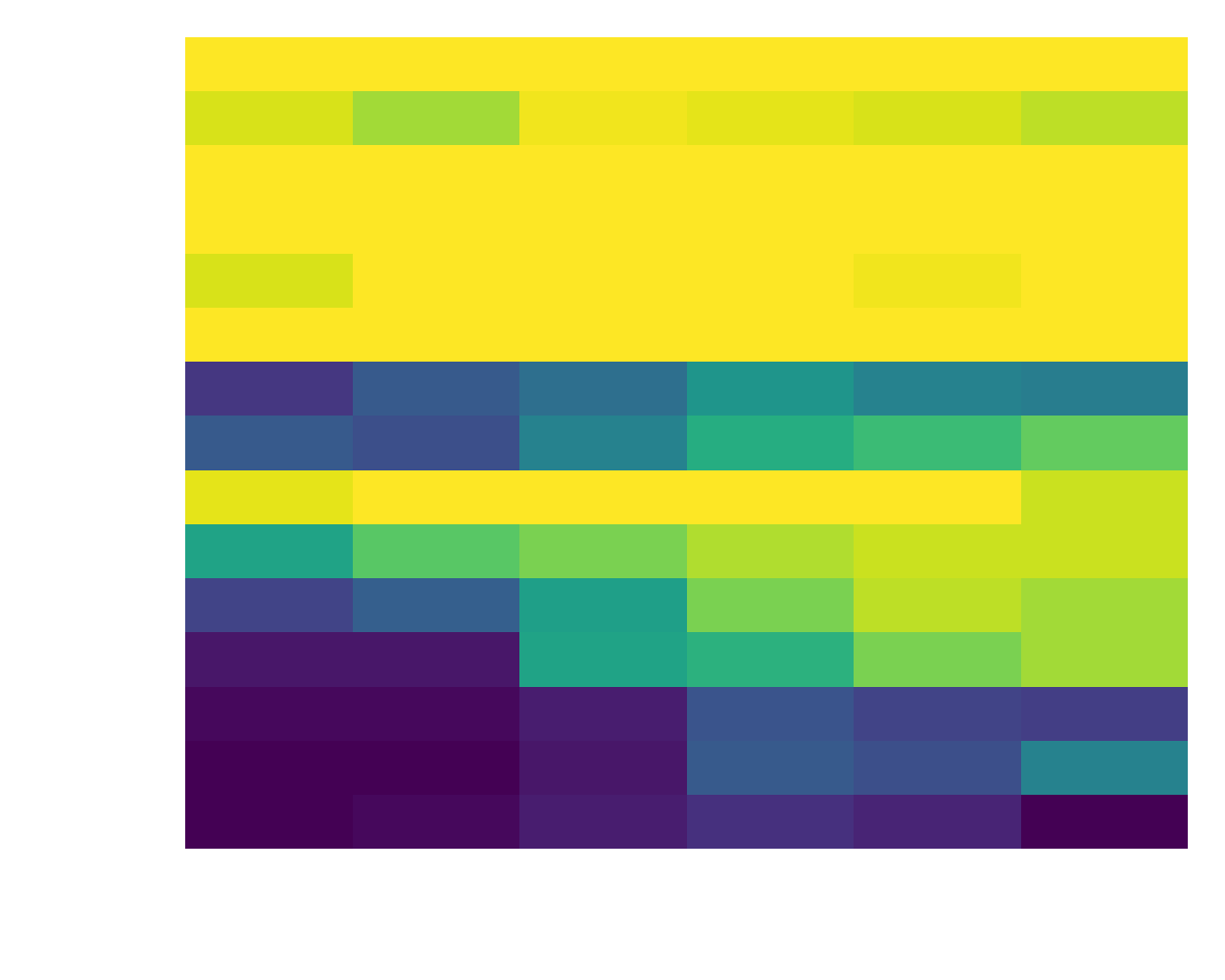
		\caption{24 Threads}
	\end{subfigure}
	\caption{Evaluation of the success rate (i.e. $1-\text{collision rate}$) for the root parallelization with similarity vote for 1, 8 and 24 threads}
	\label{fig:MCTSRootVoteG1DetailedAll}
\end{figure}

\begin{figure}
	\fns
	\begin{subfigure}{\columnwidth}
		\centering
		\def\svgwidth{\columnwidth}
		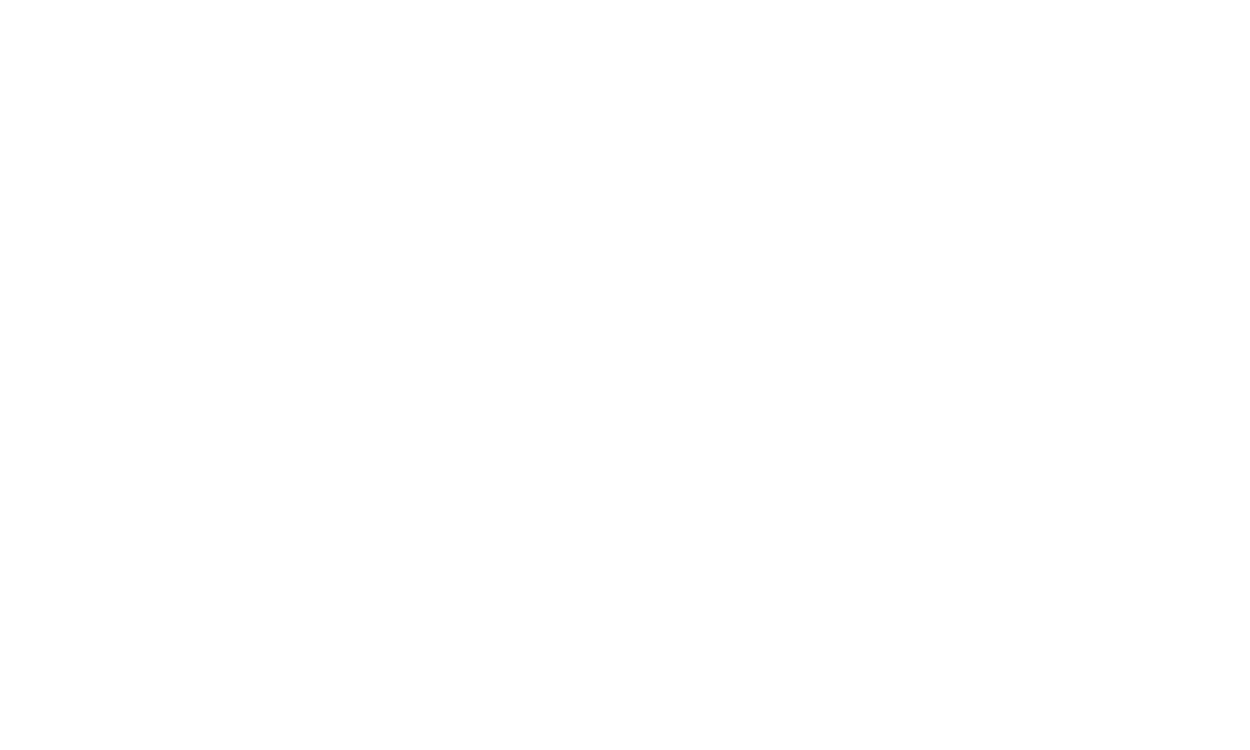
		\caption{Leaf Parallelization: Mean}
		\label{fig:MCTSLeafMean}
	\end{subfigure}
	\begin{subfigure}{\columnwidth}
		\centering
		\def\svgwidth{\columnwidth}
		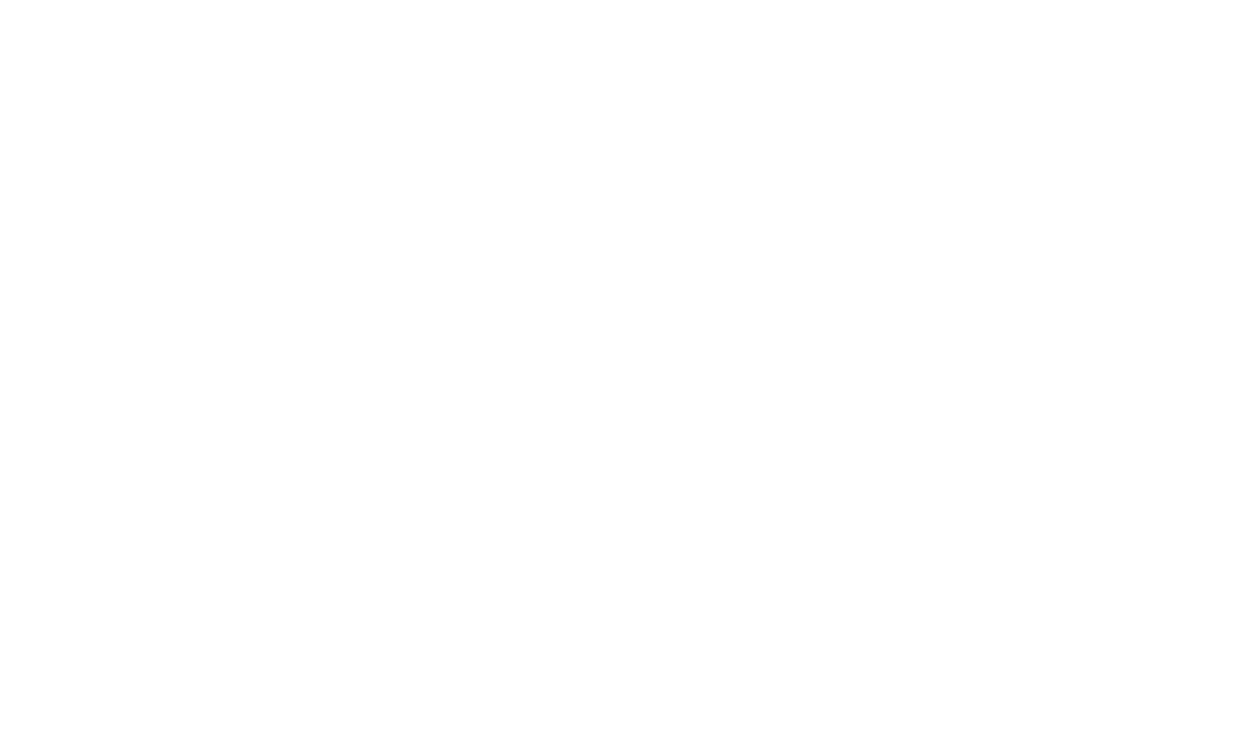
		\caption{Leaf Parallelization: Maximum}
		\label{fig:MCTSLeafMax}
	\end{subfigure}
	\begin{subfigure}{\columnwidth}
		\centering
		\def\svgwidth{\columnwidth}
		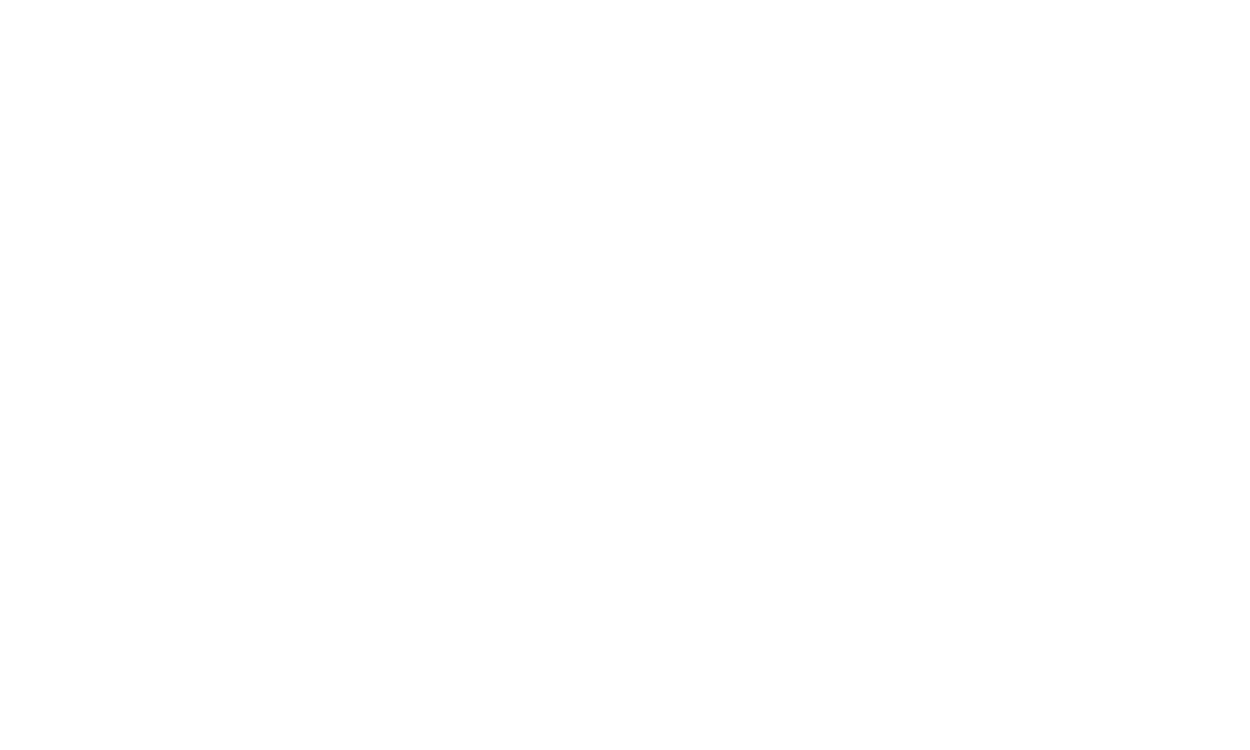
		\caption{Root Parallelization: Similarity Merge}
		\label{fig:MCTSRootMergeG5}
	\end{subfigure}
	\begin{subfigure}{\columnwidth}
		\centering
		\def\svgwidth{\columnwidth}
		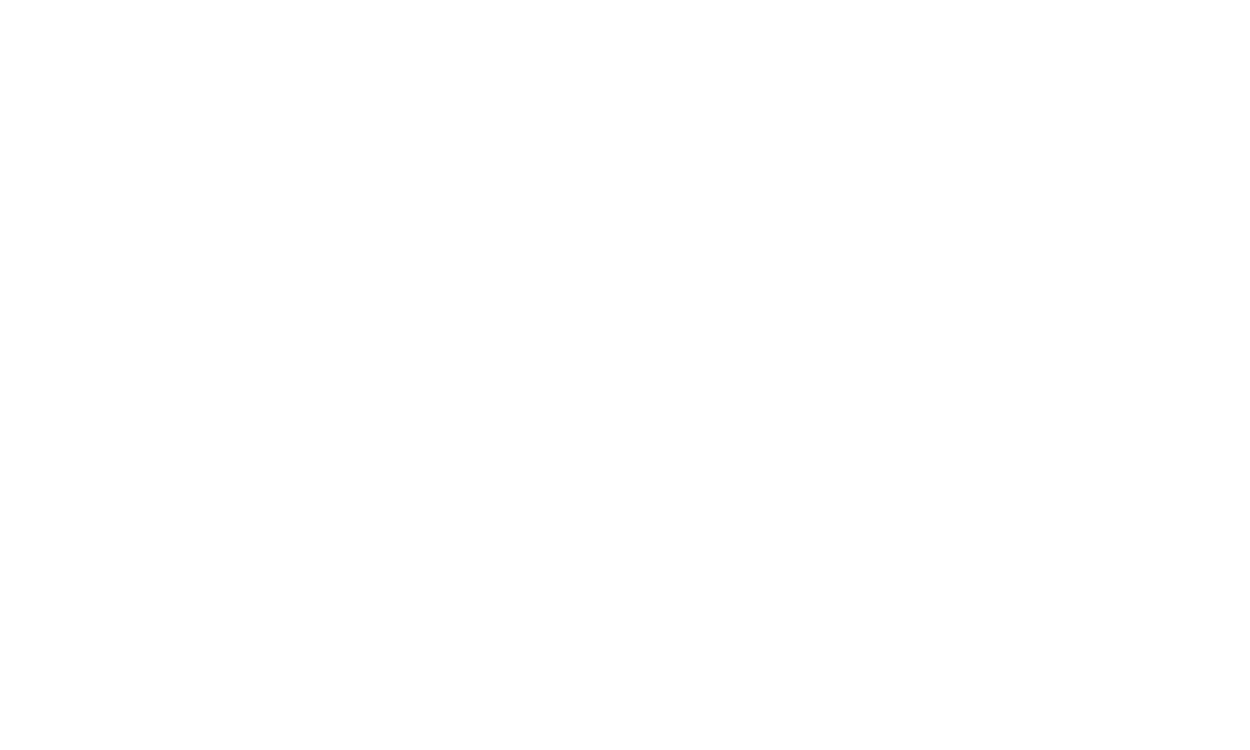
		\caption{Root Parallelization: Similarity Vote}
		\label{fig:MCTSRootVoteG1}
	\end{subfigure}
	\caption{Scalability of the different proposed parallelization strategies; The shaded region represents the deviation ($2\sigma$), of the mean success rate for the single-threaded baseline. The x-axis uses a logarithmic scale.}
	\label{fig:MCTSScalabilityAll}
\end{figure}

\appendix
For reasons of completeness the results for all scenarios are presented in Fig. \ref{fig:MCTSScalabilityAll} and Fig. \ref{fig:MCTSRootVoteG1DetailedAll}.

\bibliographystyle{IEEEtran}
\bibliography{IEEEabrv,04_mendeley-export/library}

\end{document}